\documentclass{article}

\usepackage{PRIMEarxiv}
\usepackage{amsthm,amsmath,amsfonts,amssymb}
\usepackage[authoryear]{natbib}
\usepackage{graphicx}
\usepackage{xcolor}
\usepackage{mathrsfs}
\usepackage{adjustbox}
\usepackage{algorithm}
\usepackage{algpseudocode}
\usepackage{multirow}
\usepackage{xurl}
\usepackage{pifont}

\theoremstyle{definition}

\begin{document}

\title{Poisson super learning for survival and competing risks on a discretized time scale}
\author{%
Gabriele Pittarello\\
Section of Biostatistics, Department of Public Health, University of Copenhagen\\
\texttt{gabriele.pittarello@sund.ku.dk}
\And
Thomas Gerds\\
Section of Biostatistics, Department of Public Health, University of Copenhagen\\
\texttt{tag@biostat.ku.dk}
\AND
Pernille Falberg Rønn\\
Clinical and Translational Research, Steno Diabetes Center Copenhagen,\\
Copenhagen University Hospital, Herlev, Denmark\\
\texttt{pernille.falberg.roenn@regionh.dk}
\And
Helene Charlotte Rytgaard\\
Section of Biostatistics, Department of Public Health, University of Copenhagen\\
\texttt{hely@sund.ku.dk}
}

\maketitle

\begin{abstract}
We introduce the Poisson Super Learner, a scalable ensemble method for survival and competing risks prediction with right-censored data. 
The approach exploits the correspondence between the likelihood of piecewise-constant hazard models and the Poisson likelihood. After expanding the data over a chosen time partition, candidate hazard learners are trained under a common Poisson loss, and cross-validated log-hazard predictions are combined by a Poisson meta learner. The resulting ensemble is a geometric combination of candidate hazards, and can be used to estimate survival probabilities and cause-specific absolute risks. The Poisson Super Learner framework is modular and can in principle incorporate any learner that  can estimate interval-specific hazards, including parametric, penalised, additive, and other machine-learning hazard models. Training targets the cause-specific hazard directly and does not require explicit modelling of the censoring distribution. For competing risks, separate cause-specific ensembles are fitted and subsequently combined to obtain individualised cumulative incidence predictions. We illustrate the approach using data from the Steno Diabetes Center collected between 2005 and 2022 to predict cardiovascular  disease
risk in patients with type 1 diabetes from detailed covariate information. The applied analysis is complemented by simulation studies designed to reflect key features of the data and to examine the behaviour of the method under controlled settings.
All analyses were implemented using the freely available R software package \texttt{poissonsuperlearner}.
\end{abstract}


\section{Introduction}
Individualised prediction of time-to-event outcomes usually requires
absolute risk estimates at clinically relevant time horizons. In
competing-risks settings, these risks can be obtained by combining the
cause-specific hazards for the event of interest and for the competing
events \citep{benichou90}. The statistical problem is
therefore not only to estimate flexible hazards, but also to choose
among candidate hazard estimators in a way that is targeted to
prediction.
Super Learning provides a principled framework for this problem \citep{vanderlaan07, keogh25}. Given a
library of candidate algorithms and a loss function, cross-validation is used
to construct an ensemble whose asymptotic performance is motivated by the
oracle property. For right-censored
time-to-event data, however, the choice of target and loss is central: 
survival probabilities and absolute risks are derived from hazard rates over
time, while the observed follow-up may be censored before the prediction
horizon.
In this article, we introduce the \textit{Poisson Super Learner}, a
likelihood-based ensemble method for survival and competing-risks
prediction on a discretized time scale. Follow-up is partitioned into
intervals and the cause-specific hazards are represented as piecewise
constant on this partition. Under this representation, the
counting-process likelihood is proportional to a Poisson likelihood
\citep[ch.~5]{aalen08}.
%
Piecewise exponential models can be
embedded in generalised additive models \citep{fahrmeir01,bender18} and 
include multidimensional penalised
hazard models \citep{fauvernier19} and neural-network models
parameterised on a fixed time partition \citep{kvamme2019}. 
Our contribution is different: rather than proposing
one new hazard model, we use the Poisson likelihood to combine many
candidate hazard learners under a common cross-validated criterion.
The Poisson Super Learner trains each candidate learner, collects cross-validated log-hazard
predictions, and estimates a Poisson meta learner based on the Poisson negative log-likelihood loss. The resulting
ensemble is a geometric combination of the candidate hazards. Survival
probabilities and absolute risks are then obtained as derived
quantities from the estimated cause-specific hazards. In
competing-risks problems, separate cause-specific ensembles are fitted
and combined to obtain predictions of absolute risks.
Existing Super Learner approaches for censored time-to-event data differ mainly
in the prediction target and loss function. \cite{polley11} introduced a
discrete-time Super Learner based on a longitudinal data representation in
which conditional event probabilities are estimated interval by interval using
binary outcome learners.
\cite{golmakani20} proposed
a Super Learner within the Cox proportional hazards framework, where
ensemble weights are estimated by minimising the cross-validated
negative log partial likelihood.
\cite{westling23} proposed an ensemble targeting the
entire conditional survival curve by minimising an integrated weighted Brier
score, which requires either a censoring model or coupled estimation of
survival and censoring distributions. \cite{munch24} proposed a joint survival
learner that treats censoring as an additional state and selects among tuples
of event and censoring learners. Pseudo-observation-based Super Learners use
transformed survival or cumulative-incidence outcomes for the second-stage
ensemble \citep{sachs19}, but this preprocessing can complicate monotonicity
and range restrictions for predicted risks \citep{keogh25}.
The Poisson Super Learner instead targets the cause-specific hazards
directly.  Its training criterion is the Poisson likelihood for the
piecewise-constant hazard model, so the ensemble is formed on the
hazard scale and does not require explicit modelling of the censoring
distribution. 
The contributions of this article are threefold. First, we formulate Super
Learning for discretized-time hazard regression through the
piecewise-constant hazard and Poisson likelihood representation. Second, we
propose a Poisson meta learner that combines cross-validated log-hazard
predictions, yielding a geometric ensemble of candidate hazard models. Third,
we show how the resulting cause-specific ensembles can be used to obtain
individualised survival and competing-risks predictions, and evaluate the
method in simulation studies and in an application to cardiovascular risk
prediction in patients with type 1 diabetes.
The remainder of the article is organised as follows.
Sections~\ref{sec:generalf}--\ref{sec:predictions:new} introduce the
competing-risks framework, the Poisson representation, the
cross-validation procedure, the Poisson meta learner, and prediction of
survival probabilities and cumulative incidence functions.
Section~\ref{sec:discrete:sl} introduces the discrete Poisson Super Learner, a
cross-validated selector version of the method. Section~\ref{sec:simulation}
reports simulation studies, Section~\ref{s:realdataapp} presents the Steno
Diabetes Center application, and Section~\ref{sec:conclusions} discusses
limitations and extensions.
\section{General framework}
\label{sec:generalf}
Let \(X\in\mathcal X\subseteq\mathbb R^d\) denote a vector of covariates, \(T\in(0,\infty)\) the event time, \(C\in(0,\infty)\) the censoring time, and \(\Delta\in\{1,\ldots,J\}\) the event type, where \(J\in\mathbb N\). We observe \(\widetilde T=\min(T,C)\) and \(\widetilde\Delta=\Delta I(T\leq C)\in\{0,\ldots,J\}\)
where \(\widetilde\Delta=0\) denotes censoring and
\(\widetilde\Delta=j\) denotes an observed event of type \(j\).
Our goal is to construct an ensemble estimator for the cause-specific hazard functions
$$
\lambda_j(t \mid x)
=
\lim_{h \to 0}
\frac{P\{t \le T < t+h,\ \Delta=j \mid T \ge t, X=x\}}
{h}.
$$
We assume that censoring is conditionally independent of the event time and event type given the baseline covariates.
%

%

We consider a competing risks setting with independent and identically
distributed observations
$O_i = ( X_i, \widetilde{T}_i, \widetilde{\Delta}_i)$,
$i=1,\ldots,n$. Let
\[
0=t_0<t_1<\ldots<t_m=\tau
\]
be a partition of $[0,\tau]$. 
For $k=1,\ldots,m$, define the interval indicator function 
\[
I_k(t)=I\bigl(t\in(t_{k-1},t_k]\bigr),
\]
For individual $i$ in interval $k$ we denote the event indicator for cause $j$ by
\[
D_{i,k,j}=I(\widetilde{\Delta}_i=j)\,I_k(\widetilde{T}_i),
\]
and the total time at risk by
\[
R_{i,k}
=
\max\{0,\min(\widetilde{T}_i,t_k)-t_{k-1}\}.
\]
We consider piecewise-constant approximations of the cause-specific hazard as a function of time:
\begin{equation}
\label{eq:pwchz}
\lambda_j(t\mid x)
=
\sum_{k=1}^{m} I_k(t)\lambda_{j,k}(x), 
\end{equation}
for \(x \in \mathcal{X}\). Under the piecewise-constant approximation, the
log-likelihood terms depending on \(\lambda_j\) are
\[
\ell_j(\lambda_j)(O_i)
=
\sum_{k=1}^{m}
\left[
D_{i,k,j}\log\{\lambda_{j,k}(X_i)\}
-
R_{i,k}\lambda_{j,k}(X_i)
\right].
\]
This is the log-likelihood of a Poisson model for interval-specific
event counts with mean \(  R_{i,k} \lambda_{j,k}(X_i)\).
Equivalently, for intervals with \(R_{i,k}>0\), under a log link this
corresponds to a Poisson regression model with offset \(\log(R_{i,k})\).
Intervals with \(R_{i,k}=0\) have zero likelihood contribution.
We define the corresponding loss as
the negative log-likelihood,
\begin{equation}
\label{eq:nll}
\mathscr{L}_j(\lambda_j)(O_i)
=
-\ell_j(\lambda_j)(O_i).
\end{equation}
For any function \(g\, : \, O \mapsto \mathbb{R}\), let \(P_n g = n^{-1}\sum_{i=1}^n g(O_i)\) denote the
empirical mean. The empirical risk for cause \(j\) is therefore
\[
P_n \mathscr{L}_j(\lambda_j)
=
\frac{1}{n}\sum_{i=1}^n \mathscr{L}_j(\lambda_j)(O_i).
\]
The piecewise-constant hazard approximation does not require a particular functional form of $\lambda_{j,k}(x)$, nor does it require that this function be indexed by a finite-dimensional parameter. The only requirement is that each candidate model provides interval-specific hazard estimates on the chosen time partition. This representation allows a broad range of candidate hazard estimators for $\lambda_{j,k}(x)$ that solve the minimisation problem
\begin{equation}
\label{eq:general-lkh-criterion}
\hat \lambda_{j}\in \underset{\lambda_j \in \mathcal{H}_j}{\arg\min}\left\{
P_n \mathscr{L}_j(\lambda_j)\right\},   
\end{equation}
where $\mathcal{H}_j$ denotes the class of admissible functions for
the candidate hazard estimator. This class may encode restrictions on
the functional form, smoothness, regularisation, or other
learner-specific constraints. An example is the Highly Adaptive Lasso
model described by \cite{rytgaard2022continuous}.

\subsection{Base learners}
\label{ss:baselearners}
For each cause \(j\in\{1,\ldots,J\}\), consider a library of \(S\) candidate hazard learners. Learner \(s\in\{1,\ldots,S\}\) estimates the cause-specific hazard over a learner-specific class \(\mathcal H_j^s\). Using the Poisson loss in Equation~\ref{eq:nll}, we write the corresponding estimator as
$$ \hat{\lambda}_j^s \in \arg\min_{\lambda_j\in\mathcal H_j^s} P_nL_j(\lambda_j). $$
The resulting piecewise-constant hazard estimate is
$$ \hat{\lambda}_j^s(t\mid x) = \sum_{k=1}^m I_k(t)\hat{\lambda}_{j,k}^s(x). $$
The class \(\mathcal H_j^s\) may impose learner-specific restrictions on functional form, smoothness, or regularisation. In the following subsection, we describe the learner classes used in the empirical analyses.
\subsubsection{Learners used in the empirical analyses}
For the learner families used in the empirical analyses, we parametrise the interval-specific hazard through a learner-specific function of the covariates \(f_{j,k}^s(x;\theta_j^s)\), indexed by a parameter vector \(\theta_j^s\). Specifically,
$$ \lambda_{j,k}^s(x;\theta_j^s) = \exp\{f_{j,k}^s(x;\theta_j^s)\}, $$
where \(\theta_j^s\) denotes the learner-specific parameters. The form of \(f_{j,k}^s\) and \(\theta_j^s\) depends on the learner family. 
The corresponding piecewise-constant hazard is
\[
\lambda_j^s(t\mid x;\theta_j^s)
=
\sum_{k=1}^m
I_k(t)\lambda_{j,k}^s(x;\theta_j^s).
\]
Let \(\Theta_j^s\) denote the learner-specific parameter space, so that
\[
\mathcal H_j^s
=
\left\{
\lambda_j^s(\cdot;\theta_j^s):
\theta_j^s\in\Theta_j^s
\right\}.
\]
For these parametrised learners, the minimisation problem in
Equation~\ref{eq:general-lkh-criterion} becomes
\begin{equation}
\label{eq:parametric-models-lkh-minimization}
\hat\theta_j^s
=
\underset{\theta_j^s\in\Theta_j^s}{\arg\min}
\left\{
P_n\mathscr{L}_j
\bigl(\lambda_j^s(\cdot;\theta_j^s)\bigr)
\right\}.
\end{equation}
The resulting hazard estimator is
\[
\hat\lambda_j^s(t\mid x)
=
\lambda_j^s(t\mid x;\hat\theta_j^s).
\]
For generalised linear model learners, we use the log-linear predictor
\[
f_{j,k}^s(x;\theta_j^s)
=
\gamma_{j,k}^s
+
\sum_{\ell\in\mathcal A_j}
\beta_{j,\ell}^s x_\ell,
\]
where
\[
\theta_j^s
=
(\gamma_{j,1}^s,\ldots,\gamma_{j,m}^s,\beta_j^s).
\]
The parameters \(\gamma_{j,k}^s\in\mathbb R\) represent interval-specific
baseline effects, while
\[
\beta_j^s
=
(\beta_{j,\ell}^s)_{\ell\in\mathcal A_j},
\qquad
\mathcal A_j\subseteq\{1,\ldots,d\},
\]
contains the corresponding covariate effects.
For an elastic-net-type learner \citep{glmnet}, the learner-specific parameter space takes the form
\[
\Theta_{j,\mathrm{EN}}^s(c^s)
=
\left\{
(\gamma_{j,1}^s,\ldots,\gamma_{j,m}^s,\beta_j^s):
\frac{1-\iota^s}{2}\|\beta_j^s\|_2^2
+
\iota^s\|\beta_j^s\|_1
\leq c^s
\right\},
\]
where \(c^s\in[0,\infty)\) is a hyperparameter that determines the strength of regularisation, with smaller values corresponding to stronger regularisation, and \(\iota^s\in[0,1]\) determines the mixture between ridge and lasso regularisation. The special cases \(\iota^s=1\) and
\(\iota^s=0\) correspond to lasso and ridge constraints, respectively.
For a generalised additive model learner \citep{wood17}, we partition the covariate indices
into two disjoint sets
\[
\mathcal S^s,\mathcal W^s\subseteq\{1,\ldots,d\},
\qquad
\mathcal S^s\cap\mathcal W^s=\emptyset,
\]
where \(\mathcal S^s\) contains the covariates entering through smooth effects
and \(\mathcal W^s\) contains the covariates entering linearly. The predictor
is then written as
\[
f_{j,k}^s(x;\theta_j^s)
=
\gamma_{j,k}^s
+
\sum_{\ell\in\mathcal S^s}
\zeta_{j,\ell}^s(x_\ell)
+
\sum_{\ell\in\mathcal W^s}
\beta_{j,\ell}^s x_\ell.
\]
Here, \(\gamma_{j,k}^s\in\mathbb R\) is an interval-specific baseline effect,
\(\zeta_{j,\ell}^s:\mathbb R\to\mathbb R\) denotes the smooth effect of
covariate \(x_\ell\) for cause \(j\) in learner \(s\), and
\(\beta_{j,\ell}^s\in\mathbb R\) is the linear coefficient of the
\(\ell\)-th covariate. Here, \(\theta_j^s\) collects the interval-specific baseline effects, the linear coefficients, and the spline-basis coefficients representing the smooth components.
For the GAM learner, define
\[
J_\ell(\zeta_{j,\ell}^s)
=
\int
\left\{
\frac{d^2}{dz^2}\zeta_{j,\ell}^s(z)
\right\}^2
\,dz.
\]
A constrained parameter space is then
\[
\Theta_{j,\mathrm{GAM}}^s
\left((c_\ell^s)_{\ell\in\mathcal S^s}\right)
=
\left\{
\theta_j^s\in\Theta_j^s:
J_\ell(\zeta_{j,\ell}^s)\leq c_\ell^s,\;
\ell\in\mathcal S^s
\right\},
\]
where
\(
(c_\ell^s)_{\ell\in\mathcal S^s}
\in
[0,\infty)^{|\mathcal S^s|}.
\)
In practice, the GAM learner is implemented through a penalised spline representation, where smoothing parameters control the flexibility of the fitted functions.
\section{V-Fold Cross-Validation}
\label{sec:ensemble}
In this section, we describe how out-of-sample predictions are
obtained from each candidate hazard learner using
cross-validation. These predictions are evaluated under the Poisson
loss and form the level-one data for the Poisson Super Learner \citep{breiman96-stacked}.
Within each cross-validation fold, two steps are carried out: first, the base learners are estimated on the training portion of the data; second, predictions are generated for the held-out fold. We index the folds with $v=1,\ldots, V$ and denote by
$$
\mathcal{D}_v \subset \{1,\ldots, n\}
$$
the validation set in the $v$-th fold, so that 
$$
\bigcup_{v=1}^V \mathcal{D}_v = \{1,\ldots,n\},
\qquad
\mathcal{D}_{v_1} \cap \mathcal{D}_{v_2} = \emptyset
\quad
\text{for } v_1 \neq v_2 .
$$
For each fold \(v\), we define the corresponding training set by
\[
\mathcal{T}_v=\{1,\ldots,n\}\setminus\mathcal{D}_v.
\]
The folds are defined at the subject level. Thus, all rows
corresponding to a given individual are kept together in either the training
or validation split. 
In the remainder of the manuscript, we refer to $\mathcal{T}_v$ as the $v$-th training split and to $\mathcal{D}_v$ as the $v$-th fold or validation split. The empirical distributions of the training and validation data are denoted by $P_{n,\mathcal{T}_v}$ and $P_{n,\mathcal{D}_v}$, respectively. We use \(v(i)\) to denote the  validation fold containing individual \(i\), with $v:\{1,\ldots,n\}\to\{1,\ldots,V\}$. 
\subsection{Estimating the base learners on the training folds}
For each cause \(j\in\{1,\ldots,J\}\), learner
\(s\in\{1,\ldots,S\}\), and fold \(v\in\{1,\ldots,V\}\), the base
learner is estimated using only the training observations in
\(\mathcal T_v\). We denote the corresponding fold-specific hazard
estimator by
\[
\hat\lambda_j^{s,v}
\in
\underset{\lambda_j\in\mathcal H_j^s}{\arg\min}
\left\{
P_{n,\mathcal T_v}\mathscr L_j(\lambda_j)
\right\},
\]
where \(
P_{n,\mathcal T_v}\mathscr L_j(\lambda_j)
=
\frac{1}{|\mathcal T_v|}
\sum_{i\in\mathcal T_v}
\mathscr L_j(\lambda_j)(O_i).
\)
Here, \(\mathscr L_j\) is the negative Poisson log-likelihood loss
defined in Equation~\ref{eq:nll}. Learner-specific tuning parameters are selected using only the observations in the training data \(\mathcal T_v\).
%
The resulting fold-specific piecewise-constant hazard estimate is
\[
\hat\lambda_j^{s,v}(t\mid x)
=
\sum_{k=1}^m
I_k(t)\hat\lambda_{j,k}^{s,v}(x).
\]
%
%
\subsection{Predicting on the validation folds}
For each individual \(i = 1, \ldots, n\), cause \(j=1, \ldots, J\), interval
\(k=1,\ldots,m\), and  base learner \(s=1,\ldots, S\), the learner estimated using the corresponding training set \(\mathcal{T}_{v(i)}\) is used to compute the validation-set log-hazard prediction
\[
Z_{i,j,k}^s
=
\log \hat\lambda_{j,k}^{s,v(i)}(X_i),
\]
Collecting these predictions over all learners gives the vector
\(
Z_{i,j,k}
=
\bigl(
Z_{i,j,k}^1,\ldots,Z_{i,j,k}^S
\bigr)^\top
\in\mathbb{R}^S .
\)
Repeating this procedure over all folds yields, for every individual \(i\),
cause \(j\), and interval \(k\), a cross-validated vector of log-hazard
predictions \(Z_{i,j,k}\). These predictions are out-of-sample in the sense
that \(Z_{i,j,k}\) is computed from models estimated without using observation
\(i\). We denote the level-one data for cause \(j = 1, \ldots J\)  as
\[
O_{i,j}^\star = \left\{ Z_{i,j,k}, \
D_{i,k,j},\ R_{i,k}
:
k=1,\ldots,m
\right\}.
\]
%
These data retain the interval-specific event indicators \(D_{i,k,j}\)
and times at risk \(R_{i,k}\) from the discretized representation in
Section~\ref{sec:generalf}, but differ in two respects. First, the
cross-validated log-hazard predictions from the base learners
\(Z_{i,j,k}\) replace the original covariates \(X_i\). Second, the
level-one data used to construct the Poisson meta learner are
cause-specific.
\subsection{Tuning and library construction}
\label{ss:hp-selection}
The Poisson Super Learner involves two types of tuning choices. The
first are global configuration choices, such as the number  $m$
of intervals in the time partition and the number of cross-validation
folds $V$. The second are learner-specific tuning parameters, such as
penalty parameters in regularised regression or smoothing parameters
in GAMs. When a learner has an internal tuning procedure, this tuning
is performed inside each training split of the outer
cross-validation loop. Other choices are handled by defining multiple
candidate learners. For example, lasso and ridge learners correspond
to different values of the elastic-net mixing parameter and are
treated as separate learners in the library. Computation time
increases with the number of intervals, folds, causes, and candidate
learners. We examine the sensitivity of prediction performance to the
number of intervals and folds in Web Appendix C.4. A comparison of the computation times of
the Super Learner for some fixed configurations, the base learners,
and the competing methods is provided in Web Appendix B.3.
 \begin{table}
 \caption{Hyperparameters of the Poisson Super Learner. The first block summarises the hyperparameters of the meta learner, and the remaining blocks summarise the hyperparameters of the base-learner families considered in this article, namely GLMnet, and GAM. }
 \label{tab:hp_sl}
 \begin{tabular}{@{}p{3.2cm}p{5cm}p{5cm}@{}}
 \hline
 \textbf{Input} & \textbf{Description} & \textbf{Parameter range} \\
 \hline
 \multicolumn{3}{@{}l@{}}{\textbf{Meta learner}} \\
 \hline
 Number of base-learners & $S$, number of base-learners included in the ensemble. & $\mathbb{Z}^+$ \\
 Number of folds & $V$, number of folds used in internal cross-validation. & $\mathbb{Z}^+$ \\
 Number of intervals & $m$, number of intervals in the time partition. & $\mathbb{Z}^+$  \\
 \hline
 \multicolumn{3}{@{}l@{}}{\textbf{GLMnet}} \\
 \hline
Elastic Net Mixing Parameter & $\iota$, balance between L1 (lasso) and L2 (ridge). $\iota = 1$ yields lasso; $\iota = 0$ yields ridge. & $[0, 1]$  \\
 Regularisation Path & $\xi$, value or sequence of values controlling regularisation. & $\mathbb{R}^+$  \\
 \hline
\multicolumn{3}{@{}l@{}}{\textbf{GAM}} \\
 \hline
Choice of the basis & Dimension of the basis used for smoothing. & $\mathbb{Z}^+$  \\
 \hline
\end{tabular}%
\end{table}
\section{The meta learner}
\label{sec:meta}
The purpose of the meta learner is to combine the cross-validated
out-of-sample log-hazard predictions from the base learners using the negative
Poisson likelihood loss.
%
The meta-learning step is formulated as another Poisson regression problem on
the level-one data  \( O_{i,j}^\star\) from Section~\ref{sec:ensemble}. We use a linear meta learner on the
log-hazard scale. For cause \(j\), let
\[
\alpha_j=(\alpha_j^1,\ldots,\alpha_j^S)^\top\in\mathbb R^S
\]
denote the meta learner coefficients, with one coefficient for each of the \(S\) base learners.
We use a time-invariant meta learner which uses the same coefficient \(\alpha_j\) for all time intervals.
This keeps the meta learner low-dimensional and avoids having to estimate
ensemble weights separately in each time interval.
For interval \(k\), define
\[
m_{j,k}(Z_{i,j,k};\alpha_j)
=
\exp\left\{
\sum_{s=1}^S \alpha_{j}^{s} Z_{i,j,k}^s
\right\}.
\]
The corresponding piecewise-constant meta learner hazard is
\[
m_j(t\mid Z_{i,j};\alpha_j)
=
\sum_{k=1}^m I_k(t)\,
m_{j,k}(Z_{i,j,k};\alpha_j),
\]
where \(Z_{i,j}=(Z_{i,j,1},\ldots,Z_{i,j,m})\).
Because \(Z_{i,j,k}^s=\log \hat\lambda_{j,k}^{s,v(i)}(X_i)\), the
interval-specific meta learner hazard can be written as
\[
m_{j,k}(Z_{i,j,k};\alpha_j)
=
\prod_{s=1}^S
\left\{
\hat\lambda_{j,k}^{s,v(i)}(X_i)
\right\}^{\alpha_{j}^{s}}.
\]
Thus, the Poisson meta learner combines the candidate hazards on the
hazard scale through a multiplicative combination of the base-learner hazard
estimates. If the coefficients are nonnegative, the combination has the form
of a weighted geometric product.
This differs from the standard Super Learner, which typically combines
candidate predictions through an additive (arithmetic) mixture
\citep{vanderlaan07}.
The Poisson loss applied to the level-one data is 
\[ \mathscr L_j^\star(m_j)(O_{i,j}^\star) = - \sum_{k=1}^m \left[ D_{i,k,j} \log\{m_{j,k}(Z_{i,j,k};\alpha_j)\} - R_{i,k} m_{j,k}(Z_{i,j,k};\alpha_j) \right]. \] 
This has the same form as the Poisson loss used to fit the base learners, with the original covariates replaced by the cross-validated log-hazard predictions. For each cause \(j\), the coefficient vector is estimated by minimising the empirical Poisson loss on the level-one data: 
\[ \hat\alpha_j = \arg\min_{\alpha_j\in\mathbb R^S} P^*_n \mathscr L_j^\star(m_j(\cdot; \alpha_j)), \] 
where \( P^*_n \mathscr L_j^\star(m_j) = \frac{1}{n} \sum_{i=1}^n \mathscr L_j^\star(m_j)(O_{i,j}^\star). \) The estimated meta learner hazard for cause \(j\) is \[ \hat m_j(t\mid Z_{i,j}) = \sum_{k=1}^m I_k(t)\, \hat m_{j,k}(Z_{i,j,k}), \] with \( \hat m_{j,k}(Z_{i,j,k}) = m_{j,k}(Z_{i,j,k};\hat\alpha_j). \)
\section{The Discrete Poisson Super Learner} 
\label{sec:discrete:sl}
The previous sections define an ensemble that combines candidate hazard
learners through a Poisson meta learner. In this section, we introduce the discrete Poisson Super Learner. For each competing risk \(j \in \{1, \ldots J\} \), it selects the single learner \(s  \in \{ 1, \ldots, S\}\) from the library that minimises the cross-validated Poisson loss across the $V$-folds. Let \(\hat\lambda_{j,k}^{s,v}\) denote the  hazard estimate
produced by learner \(s\) using the corresponding training fold. The discrete Poisson Super Learner for each cause \(j\), selects
\[
\hat s_j
=
\underset{s\in\{1,\ldots,S\}}{\arg\min}
\left\{
\frac{1}{n}
\sum_{v=1}^V
\sum_{i\in\mathcal D_v}
\mathscr L_j\!\left(\hat\lambda_j^{s,v}\right)(O_i)
\right\}.
\]
%
The selected learner for cause \(j\) is then re-estimated using all $n$ observations and used for prediction. 
\section{Predictions on new data}
\label{sec:predictions:new}
%

Let $\hat\lambda_j^s$ be the full-sample estimate of base learner \(s\) for cause \(j\). For a covariate vector \(x^{\mathrm{new}}\in\mathcal X\), compute the log-hazard
prediction from each base learner, for \(s=1,\ldots,S\) and
\( k=1,\ldots,m\),
\[
Z_{j,k}^{s,\mathrm{new}}
=
\log \hat\lambda_{j,k}^s(x^{new}),
\]
with \(
Z_{j,k}^{\mathrm{new}}
=
\left(
Z_{j,k}^{1,\mathrm{new}},\ldots,
Z_{j,k}^{S,\mathrm{new}}
\right)^\top
\). The ensemble cause-specific hazard is then
\[
\hat\lambda_j(t\mid x^{new})
=
\sum_{k=1}^m I_k(t)\hat\lambda_{j,k}(x^{new}),
\]
where
\[
\hat\lambda_{j,k}(x^{new})
=
m_{j,k}(Z_{j,k}^{\mathrm{new}};\hat\alpha_j)
=
\exp\left\{
\sum_{s=1}^S
\hat\alpha_{j}^{s}
Z_{j,k}^{s,\mathrm{new}}
\right\}.
\]
%
%
%
\subsection{Risk predictions}
\label{sec:riskpred}
The cause-specific hazard estimates are mapped to estimated survival
probabilities and estimated absolute risks using well-known
formulae \citep{benichou90,abgk-book}.
Let
\[
\hat\Lambda_j(t\mid x)
=
\int_0^t \hat\lambda_j(u\mid x)\,du
\]
denote the estimated cumulative cause-specific hazard, and define the survival function
\[
\hat S(t\mid x)
=
\exp\left\{
-\sum_{\ell=1}^J \hat\Lambda_\ell(t\mid x)
\right\}.
\]
The absolute risk for cause \(j\) is
\[
F_j(t\mid x)
=
\int_0^t S(u-\mid x)\lambda_j(u\mid x)\,du.
\]
For piecewise-constant hazards, this integral has an explicit form (arguments provided in the Supplementary Material). Let
\[
\hat\lambda_{\cdot,k}(x)
=
\sum_{\ell=1}^J \hat\lambda_{\ell,k}(x)
\]
denote the estimated total hazard in interval \(k\), and define
\[
\Delta_k(t)
=
\max \left\{0, \min(t,t_k)-t_{k-1}\right\}.
\]
Then the estimated absolute risk of an event of cause \(j\) by time \(t\) is 
\[
\hat F_j(t\mid x)
=
\sum_{k=1}^{m}
\hat S(t_{k-1}\mid x)
\frac{\hat\lambda_{j,k}(x)}
{\hat\lambda_{\cdot,k}(x)}
\left[
1-
\exp\left\{
-\hat\lambda_{\cdot,k}(x)\Delta_k(t)
\right\}
\right],
\]
with the convention that \(0 / 0 = 0\).
\section{Simulation study}
\label{sec:simulation}
We examine key aspects of the Poisson Super Learner. The study is
divided into two parts. The first part focuses on the right-censored
survival setting where we have a common basis for comparison with
existing benchmark implementations. In this part, we compare the
behaviour of the proposed approach with alternative super learning
methods. We focus on three main questions. First, we assess whether
the Poisson Super Learner performs similar as the best candidate
learners included in its library. This corresponds to the oracle
property discussed in \cite{keogh25}, namely the ability of the Super
Learner to perform at least as well as the best learner in the
candidate library. In settings where the library contains a learner
that is expected to approximate the data-generating process well, we
also examine whether the Discrete Super Learner selects this
learner. Second, we study how the composition of the
learner library affects the prediction performance by comparing a
reduced linear library with a richer library that additionally
includes a nonlinear additive learner. This comparison is carried out
under two simulated scenarios, corresponding respectively to linear
and nonlinear structures in the true hazard. Third, we compare the
proposed method with existing survival super learning approaches and a
proportional-hazards benchmark. We repeat these comparisons across
several training sample sizes in order to assess finite-sample
behaviour, while evaluating all models on the same large
independent test sample. In the second part, we examine the
scalability of the proposed approach for large data
sets. Here, we simulate five data sets from a controlled
competing-risks scenario in which the true cause-specific hazard
satisfies the assumptions of a standard cause-specific Cox model. We
then compare the computation time of the Poisson Super Learner with
that of the cause-specific Cox model. This analysis is used to assess
how the relative computational cost of the proposed approach changes
as the sample size increases, thereby illustrating its potential for
large-sample applications such as analyses of national registry data.
\subsection{Data-generating mechanisms}
The data-generating mechanisms were chosen to reflect key features of the Steno type 1 diabetes cohort analysed in Section \ref{s:realdataapp}, while allowing the true signal structure to be controlled. The simulator generates hypothetical patients with baseline covariates motivated by the real application and dependence patterns induced through a structural equation model \citep{holst13}. We consider two scenarios that share the same baseline covariate distribution, censoring mechanism, and competing-risk hazard, but differ in how the cause 1 event hazard is generated:
\begin{itemize}
\item In \textit{Scenario Alpha}, the cause 1 hazard depends linearly on all covariates.
\item In \textit{Scenario Beta}, the linear effects of age and LDL cholesterol on the cause 1 hazard are supplemented by nonlinear hinge-squared terms.
\end{itemize}
The comparison between the two settings therefore isolates the effect of additional flexibility in the learner library when the cause 1 hazard departs from linearity.
\subsection{Poisson Super Learner variants and component learners}
\label{sec:psl-methods-compared}
We compare two Poisson Super Learner ensembles, the discrete version,
and the individual learners entering the Poisson Super Learner library. This
allows us to separate the effect of library composition from the effect of
hazard-scale ensemble construction. All methods are fitted using the same
baseline covariates and the same time partition. The full learner library consists of four learners: an unpenalised generalised
linear model (\textbf{GLM}), a lasso-penalised generalised linear model
(\textbf{Lasso}), a ridge-penalised generalised linear model (\textbf{Ridge}),
and a generalised additive model (\textbf{GAM}). The Poisson Super Learner
using this library is denoted \textbf{PSL-Full}. The corresponding discrete
version, which chooses the single learner with the lowest cross-validated
Poisson loss, is denoted \textbf{Discrete-PSL}. We also consider a reduced
linear library containing only \textbf{GLM}, \textbf{Lasso}, and
\textbf{Ridge}; the Poisson Super Learner using this reduced library is denoted
\textbf{PSL-Linear}. Thus, \textbf{PSL-Full} and \textbf{PSL-Linear} differ
only in whether the nonlinear \textbf{GAM} learner is included. Comparing the Poisson Super Learner with a richer learner library to its reduced version based only on generalised linear learners then shows when additional flexibility is beneficial and when a simpler library is sufficient. 
\subsection{External benchmark methods}
\label{sec:competitors}
We compare the Poisson Super Learner variants with four external benchmark
methods for right-censored survival prediction. The benchmarks were chosen to
represent both standard survival modelling and recently proposed ensemble
approaches for time-to-event outcomes. Throughout the simulation results, we
use the labels \textbf{Cox}, \textbf{JoSL}, \textbf{survSL}, and
\textbf{Pseudo SL} for these methods. 

\textbf{Cox}. As a standard regression benchmark, we fit a Cox proportional hazards model.
The Cox model was specified using knowledge of the simulation design and should
therefore be interpreted as a strong design-informed benchmark. In
Scenario Alpha, the model includes the baseline covariates as linear terms. In
Scenario Beta, the model additionally includes the nonlinear age and LDL
cholesterol terms used in the event-generating mechanism. Thus, the Cox
benchmark is deliberately favourable, especially with respect to the
scenario-defining effects.

\textbf{JoSL}.
We include the joint survival learner of \cite{munch24}, denoted
\textbf{JoSL}. This method evaluates combinations of learners for the event
and censoring mechanisms. In our implementation, the learner library consists
of an intercept-only model, a Cox proportional hazards model, and a penalised
Cox model for both the event and censoring components. Because this library is
based primarily on linear and penalised regression structures, we expect
\textbf{JoSL} to behave similarly to methods based on the reduced linear
library, especially in Scenario Alpha.

\textbf{survSL}.
We also include the continuous-time survival Super Learner of
\cite{westling23}, denoted \textbf{survSL}. This method targets the
conditional survival curve and uses an integrated Brier-score criterion. We
use Kaplan--Meier, Cox, and generalised additive model learners for both the
event and censoring models, with three cross-validation folds.

\textbf{Pseudo SL}.
Finally, we include a pseudo-observation-based Super Learner, denoted
\textbf{Pseudo SL}, based on \cite{sachs19}. This approach provides a
conceptually different route to ensemble prediction: pseudo-observations
derived from nonparametric survival or cumulative-incidence estimators are used
as outcomes in a second-stage prediction model. In our implementation, the
learner library consists of a generalised linear model, a penalised generalised
linear model, and a generalised additive model. For prediction at five years,
pseudo-observations are computed on the grid \(\{1,\ldots,10\}\).
\subsection{Performance measure}
All methods are evaluated at the same prediction horizon of 5 years and on the same large test sample within each simulated scenario, using the Index of Prediction Accuracy \citep[IPA,][]{gerds21} as performance measure. The IPA is a scaled Brier-score measure. It compares the prediction error of a
model with that of a null model that assigns the same risk to all individuals.
An IPA of 0 corresponds to the null model, while larger values indicate better
prediction accuracy.
In formula, the Index of Prediction Accuracy (IPA) for cause \(j\) is defined as
\[
\text{IPA}_j(t)
=
1-
\frac{\mathscr B_{j,t}(\hat F_j)}
{\mathscr B_{j,t}(\hat F_{j,0})},
\]
where $\mathscr B_{j,t}(\hat F_j)$ is the Brier score for the time-to-event model being evaluated, while $\mathscr B_{j,t}(\hat F_{j,0})$ is the Brier score for a reference model, here the null model with no covariates.
In right-censored settings, as in the real data application, the Brier score can be calculated using inverse probability of censoring weighting (IPCW). Because the uncensored event times were available in the simulation, prediction accuracy was evaluated without estimating the censoring distribution. This isolates prediction performance from additional variability due to IPCW estimation.
Specifically, each fitted model was evaluated at \(t=5\) on an independently simulated test sample using the binary outcome $Y_{i,j}^{\ast} = I\!\left(T_i \leq 5,\Delta_i=j\right)$. The Brier score of a fitted model was calculated as
\[
\mathscr{B}_{j,5}^{\ast}(\hat F_j)
=
\frac{1}{N}
\sum_{i=1}^{N}
\left[
Y_{i,j}^{\ast}-\hat F_j(5 \mid X_i)
\right]^2.
\]
The null model assigned every subject the same absolute-risk prediction, equal to the empirical proportion of subjects experiencing event \(j\) by time \(5\) in the uncensored test sample: $\hat F_{j,0}(5) = \frac{1}{N}
\sum_{i=1}^{N}
Y_{i,j}^{\ast}.$ 
\subsection{Results}
\begin{figure}
\centering
\includegraphics[width=\textwidth]{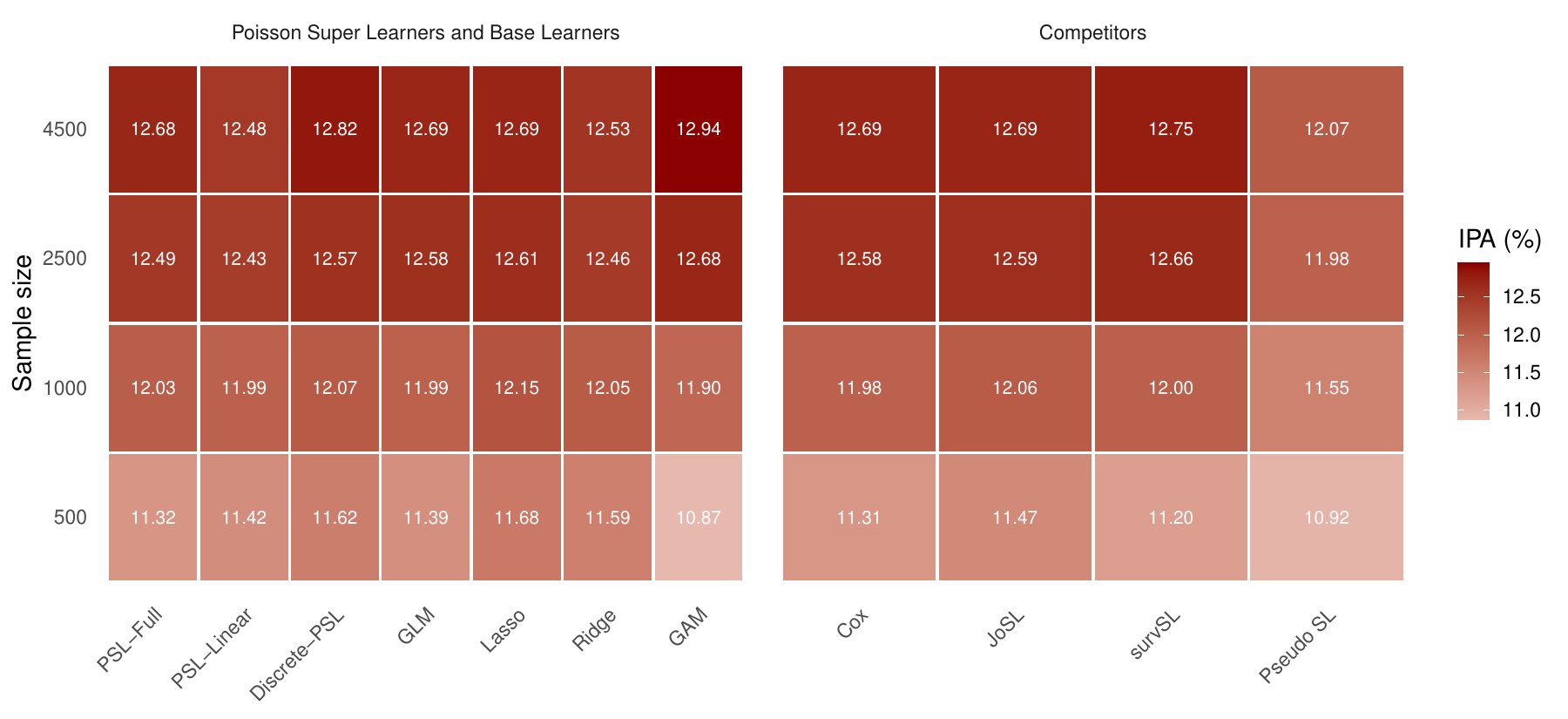}
\caption{Scenario Alpha. Average IPA scores for increasing sample sizes
(y-axis) for the different methods (x-axis).}
\label{fig:tiles-alpha}
\end{figure}

\begin{figure}
\centering
\includegraphics[width=\textwidth]{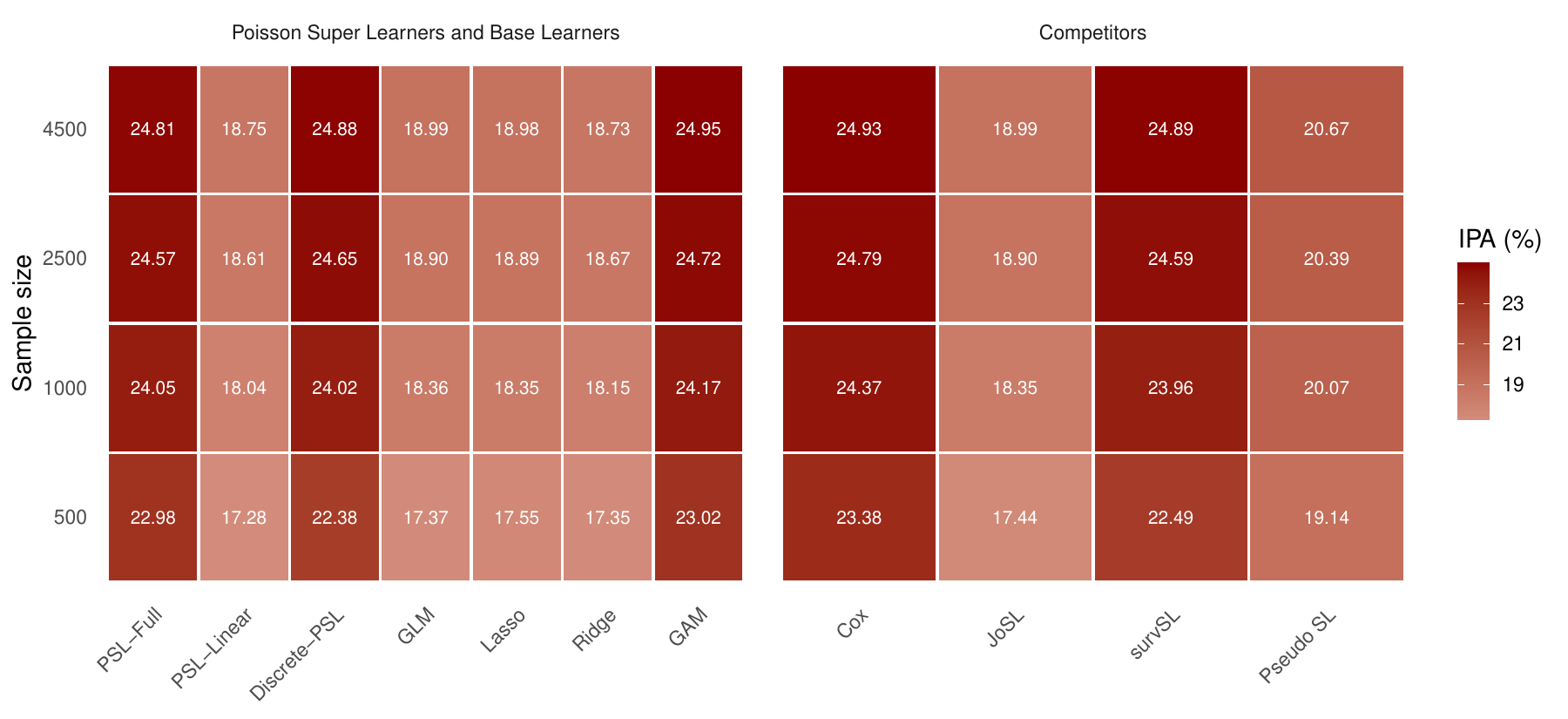}
\caption{Scenario Beta. Average IPA scores for increasing sample sizes
(y-axis) for the different methods (x-axis).}
\label{fig:tiles-beta}
\end{figure}

Figures~\ref{fig:tiles-alpha} and~\ref{fig:tiles-beta} show the average IPA scores across  80 simulation replicates for Scenarios Alpha and Beta, respectively. Rows correspond to the training sample sizes and columns to the methods under comparison. Darker colours indicate higher IPA values and hence better predictive performance. In each figure, the left panel contains the Poisson Super Learner variants and their component learners, while the right panel contains the external benchmark methods. Across both scenarios, prediction performance generally improved with increasing training sample size. Since all methods were evaluated on the same fixed large test sample within each scenario ($10000$), this pattern reflects the increased amount of information available for model fitting rather than changes in the evaluation distribution. The sample size trend should therefore be viewed primarily as a finite-sample diagnostic. The main comparisons concern how the Poisson Super Learner behaves relative to its component learners, how the learner library affects performance, and how the proposed method compares with external benchmarks. In Scenario Alpha, where the event hazard is approximately linear in the covariates, \textbf{PSL-Linear} and \textbf{PSL-Full} performed similarly. This is expected because the reduced linear library already contains learners well aligned with the data-generating mechanism. The individual \textbf{GLM}, \textbf{Lasso}, and \textbf{Ridge} learners also performed similarly, and the Poisson Super Learner variants tracked these strong library members. Thus, adding the nonlinear \textbf{GAM} learner did not materially improve performance in the approximately linear scenario, but it also did not substantially degrade performance. The external benchmarks showed a similar pattern in Scenario Alpha. \textbf{Cox} performed well, as expected, because it was specified using
knowledge of the approximately linear event-generating mechanism. \textbf{JoSL} also performed competitively, consistent with its use of linear and penalised Cox-type learners. The \textbf{survSL} benchmark performed well, while \textbf{Pseudo SL} was less competitive. Overall, in the approximately linear setting, the proposed Poisson Super Learner was competitive with the strongest benchmark methods and behaved similarly to the strongest learners in its own library. In Scenario Beta, where the event hazard included nonlinear effects of age and LDL cholesterol, the methods capable of representing nonlinear structure performed best. Within the Poisson Super Learner family, \textbf{PSL-Full} clearly outperformed \textbf{PSL-Linear} across the investigated sample sizes. This difference was driven by the inclusion of the \textbf{GAM} learner, which was well matched to the nonlinear signal and was the strongest individual learner in the full library. The performance of \textbf{PSL-Full} was close to that of \textbf{GAM}, showing that the ensemble adapted to the strongest available nonlinear learner. The \textbf{Discrete-PSL} also performed close to the strongest full-library learners, particularly at larger sample sizes. Among the external benchmarks in Scenario Beta, \textbf{Cox} remained a strong competitor because it included the nonlinear terms used in the data-generating mechanism. The \textbf{survSL} method also performed well. In contrast, \textbf{JoSL} and \textbf{Pseudo SL} improved with increasing sample size but remained below the best-performing methods in this scenario. These results suggest that the richer Poisson Super Learner library is beneficial when the true hazard structure departs from linearity, provided that the library contains learners capable of capturing the relevant nonlinear effects. Taken together, the simulation study supports two conclusions. First, the results of the Poisson Super Learner were consistent with the oracle motivation of Super Learning: the ensemble performed similarly to the strongest candidate learner across the investigated settings. Second, learner-library composition mattered: the reduced linear library was sufficient in the approximately linear scenario, whereas the full library was preferable when the hazard contained nonlinear effects. 
\subsection{Results on the scalability of the Poisson Super Learner}
\label{ss:scalability-study}
\begin{figure}
\centering
\includegraphics[width=.8\textwidth]{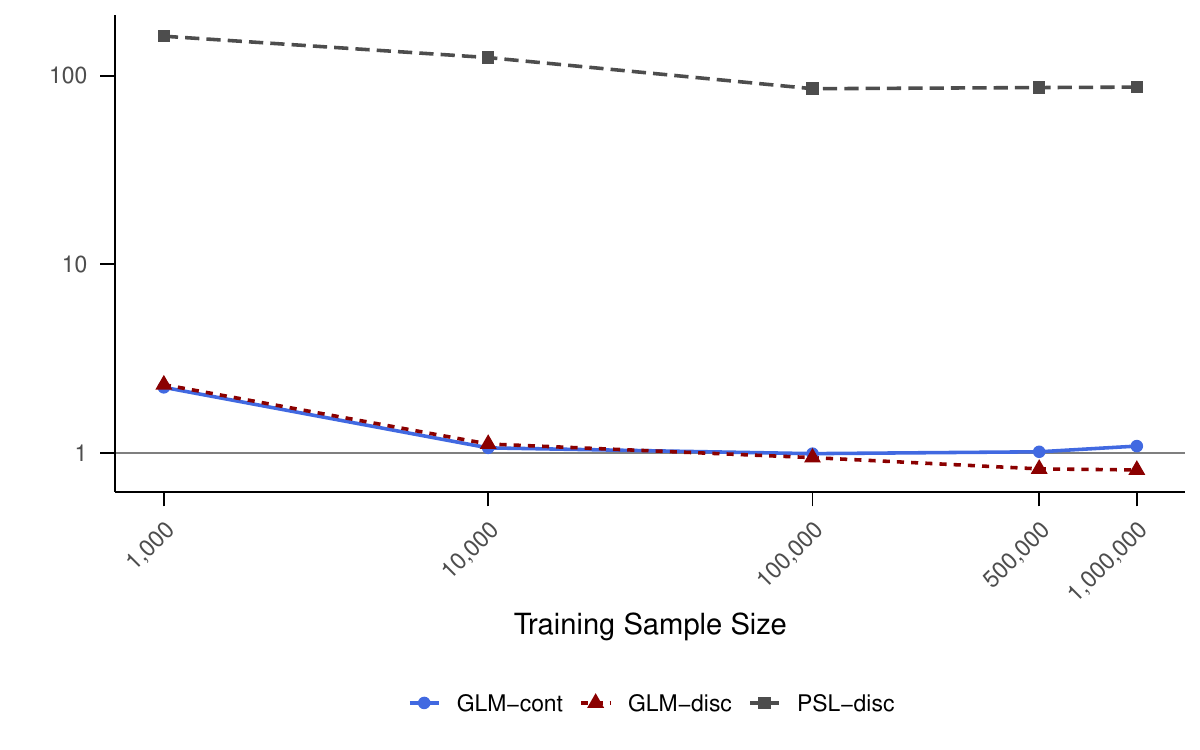}
\caption{\label{fig:relative-runtime}
Average computation time relative to the cause-specific Cox model,
\textbf{CSC-Cox}, by training sample size, based on five data sets from
Scenario Alpha. Average computation times were calculated over ten refits.
The horizontal line at \(y=1\) denotes the \textbf{CSC-Cox} benchmark.
The Poisson-based models are \textbf{PSL-disc}, \textbf{GLM-cont}, and
\textbf{GLM-disc}.}
\end{figure}

Piecewise-constant hazard models are computationally attractive in large data sets with categorical covariates, since observations sharing the same covariates can be aggregated. We refer to this well-known property as the computational scalability of Poisson regression models; see Web Appendix B.4 and \cite[p.~225]{aalen08}. In this section, we assess the computational scalability of the proposed approach empirically in a controlled competing-risks setting. We simulate data from scenario Alpha with sample sizes $n = 1000$, $10000$, $100000$, $500000$, and $1000000$. To demonstrate our point, we compare four models. The first, \textbf{CSC-Cox}, is a cause-specific Cox model fitted to the synthetic version of the original covariates listed in Table \ref{tab:data}. It satisfies the assumptions of the data generator and serves as a correctly specified benchmark. We fit it using \texttt{coxph} from the \texttt{R} package \texttt{riskRegression} \citep{ozenne17}. The second, \textbf{GLM-cont}, is an unpenalised Poisson generalised linear model using the same covariates. The third, \textbf{GLM-disc}, uses the same model after discretising the numeric covariates by rounding them down to the nearest integer using \texttt{floor}. The fourth model is the Poisson Super Learner, also fitted to the discretised covariates. For the cause of interest, its library comprises unpenalised, lasso-penalised, and ridge-penalised Poisson generalised linear models; for the competing event, it comprises unpenalised and lasso-penalised models. Penalty parameters are selected by internal cross-validation. The Super Learner uses $V=3$ and $3$ time knots. Figure \ref{fig:relative-runtime} shows the average runtime of each fitted model relative to \textbf{CSC-Cox} as the training sample size increases, computed over $10$ refits on the same data. The choice of $10$ refits was based on the experiment reported in Web Appendix B.5, where the benchmark model was refitted $30$ times to data from scenario Alpha with sample size $100000$. In that experiment, the variability of the average runtime was negligible after $10$ refits. The reference runtime of \textbf{CSC-Cox} is represented by the horizontal line at $y=1$. The relative runtime decreases with increasing sample size for the Poisson-based models. Comparing \textbf{GLM-cont} and \textbf{GLM-disc}, the results show that discretising the covariates reduces computation time. This reduction is most pronounced for large sample sizes, where \textbf{GLM-disc} is faster than the \textbf{CSC-Cox} benchmark. Absolute runtimes in seconds are provided in Web Table 3.
Predictive performance was essentially unchanged: across all sample sizes, the Poisson-based models achieved Brier scores comparable to those of the correctly specified \textbf{CSC-Cox} benchmark; see Web Figure 4. The simulation study presented in this section further motivates the Poisson Super Learner. In the spirit of piecewise-constant regression, discretising the covariates allows the proposed approach to use computationally inexpensive model configurations, improving computational efficiency relative to approaches based on the partial likelihood without materially affecting predictive performance.
\section{Data application on the real data}
\label{s:realdataapp}
In this Section, we apply our methods to the data set of 4821 adult patients with type 1 diabetes form Steno Diabetes Center Copenhagen, Herlev, Denmark. The data were collected between 2005 and 2022 from electronic patient records and linked with national register data through a personal identification number to get information on diagnosis codes, death and emigration during follow-up until 30 April, 2022. We split the data into a training set comprising 3583 observations from the calendar years 2005 to 2014 and a test set comprising 1238 observations from the calendar years 2015 to 2022. Our aim is to study the time to the first occurrence of cardiovascular disease in the presence of a competing event, namely non-CVD death, using a set of individual measurements as baseline covariates. Patients with type 1 diabetes have an increased risk of developing cardiovascular disease, and this risk is influenced by their individual characteristics. The baseline covariates are summarised in Table \ref{tab:data}. The second column provides a description of each covariate. The covariates include clinician-assessed measurements, such as albuminuria status and estimated glomerular filtration rate; biological characteristics, such as age; and lifestyle factors, such as smoking behavior. For categorical covariates, Table \ref{tab:data} reports the different levels (column 3) and their distributions in the training and test sets (columns 4 and 5). For continuous covariates, the table reports the median and interquartile range in the training and test sets (columns 4 and 5).
\begin{table}
\caption{\label{tab:data} Baseline covariates of the Steno data set of patients with type 1 diabetes. The covariates are listed in column 1 and described in column 2. Column 3 reports the levels of the categorical covariates. Columns 4 and 5 report the distribution of the categorical covariates in the training and test sets, respectively, or the median and interquartile range in brackets for the continuous covariates. }
\resizebox{\textwidth}{!}{%
\begin{tabular}{@{}p{3.5cm}p{3cm}p{2cm}p{3cm}p{3cm}@{}}
\hline
\textbf{Variable}& \textbf{Description}  & \textbf{Level} & \textbf{Train: 2005-2014 (n=3583)} & \textbf{Test: 2015-2022 (n=1238)}\\
\hline
\texttt{sex} & Gender  & Woman & 1653 (46.1) & 577 (46.6)\\
\texttt{age} & Age  &  & 43.5 [31.6, 55.4] & 31.7 [22.1, 48.1]\\
\texttt{diabetes\_duration}& Duration of diabetes (years) & & 19.7 [10.6, 31.1] & 10.6 [ 3.8, 20.8]\\
\texttt{value\_SBP} &  Systolic blood pressure (mmHg) & & 131 [120.0, 143.5] & 124 [116, 133]\\
\texttt{value\_LDL}& LDL cholesterol (mmol/L)&  & 2.4 [1.9, 2.9] & 2.3 [1.9, 2.9]\\
\texttt{value\_HBA1C}&HbA1c (mmol/mol) &  & 64 [56, 74] & 60 [52, 70]\\
\texttt{value\_Smoking} &Smoking habit & Yes & 1,042 (29.1) & 292 (23.6)\\
\texttt{value\_Motion}& Physical activity & Yes & 2,472 (69.0) & 846 (68.3)\\
\texttt{value\_Albuminuria} & Albuminuria status & Normal & 2,939 (82.0) & 1,106 (89.3)\\
 && Micro & 485 (13.5) & 100 (8.1)\\
 && Macro & 159 (4.4) & 32 (2.6)\\
\texttt{eGFR} & Estimated glomerular filtration rate (mL/min/1.73m\textsuperscript{2}) &  & 107 [ 93.4, 118.3] & 114.5 [102.0, 126.9]\\
\hline
\end{tabular}}
\end{table}
The Poisson Superlearner and the Discrete Poisson Superlearner considered in this section are constructed using cause-specific libraries of learners. For cause 1, we use the same library of learners as described in Section \ref{sec:simulation} for \textbf{PSL-Linear}. For cause 2, we consider a library comprising two learners: an unpenalised generalised linear model including the covariates \texttt{sex} and \texttt{age}, and a generalised additive model in which the continuous covariate \texttt{age} is modelled using a spline term. This choice is motivated by the small number of subjects experiencing the competing event and by considerations of computational efficiency. We refer to the Poisson Superlearner as \textbf{PSL} and to the Discrete Poisson Superlearner as \textbf{Discrete-PSL}. Table \ref{tab:ipa-by-time} reports the IPA scores by prediction horizon and model. The \textbf{PSL} and the \textbf{Discrete-PSL} are compared with the Cause-Specific Cox model (\textbf{CSC}) and with four individual candidate models: \textbf{GLM}, \textbf{GAM}, \textbf{Lasso}, and \textbf{Ridge}. Each of these individual candidate models is fitted using all covariates for both causes. By contrast, the \textbf{Discrete-PSL} is constructed through a cause-specific selection procedure: for each cause, it selects the learner with the lowest cross-validated Poisson deviance from the corresponding learner library described above. For each prediction horizon, the score of the best-performing model is highlighted in bold. The \textbf{Ridge} model performs best at prediction horizons 1, 4, and 5. The \textbf{GAM} performs best at prediction horizons 6 and 7, while the \textbf{CSC} model performs best at prediction horizons 2 and 3. Although the \textbf{PSL} does not achieve the highest score, its performance is close to that of the best-performing individual model at different prediction horizons, namely \textbf{Ridge} at earlier horizons and \textbf{GAM} at later horizons. The \textbf{Discrete-PSL} performs worse than the \textbf{PSL}.
\begin{table}[htbp]
\centering
\caption{\label{tab:ipa-by-time}
Integrated prediction accuracy (IPA, \%) by prediction horizon and model.
For each prediction horizon, the highest IPA value is shown in bold.}
\begin{tabular}{@{}p{3.4cm}rrrrrrr@{}}
\hline
\textbf{Model} & \multicolumn{7}{c}{\textbf{Prediction horizon}} \\
\cline{2-8}
& \textbf{1} & \textbf{2} & \textbf{3} & \textbf{4} & \textbf{5} & \textbf{6} & \textbf{7} \\
\hline
\textbf{Null model}    & 0.0          & 0.0          & 0.0          & 0.0          & 0.0          & 0.0           & 0.0 \\
\textbf{GLM}           & 2.6          & 3.5          & 5.7          & 8.5          & 7.0          & 9.7           & 8.1 \\
\textbf{GAM}           & 1.9          & 3.3          & 5.7          & 8.0          & 7.1          & \textbf{11.0} & \textbf{9.3} \\
\textbf{Lasso}         & 2.5          & 3.5          & 5.7          & 8.5          & 7.1          & 9.7           & 8.1 \\
\textbf{Ridge}         & \textbf{2.9} & 3.7          & 6.0          & \textbf{8.7} & \textbf{7.4} & 10.1          & 8.3 \\
\textbf{PSL}           & 2.3          & 2.8          & 6.0          & 8.3          & 7.2          & 10.3          & 8.2 \\
\textbf{Discrete-PSL}  & 0.9          & 1.8          & 5.4          & 7.6          & 6.1          & 9.4           & 7.6 \\
\textbf{CSC}           & 2.8          & \textbf{3.8} & \textbf{6.2} & 8.5          & 6.9          & 9.7           & 8.2 \\
\hline
\end{tabular}
\end{table}
\section{Discussion and conclusions}
\label{sec:conclusions}
%

We introduced the Poisson Super Learner, a Super Learner framework for cause-specific hazard estimation on a discretized time scale. Across the investigated settings, the Poisson Super Learner performed similarly to the strongest candidate learners in its library. The results also showed that library composition matters: a reduced linear library was sufficient when the true hazard structure was approximately linear, whereas a richer library including a nonlinear additive learner improved performance when the event hazard contained nonlinear effects. Compared with existing survival Super Learner approaches and proportional-hazards benchmarks, the proposed method was competitive while retaining a likelihood-based interpretation on the hazard scale. The scalability experiment showed that the relative computational cost of the Poisson-based models decreased with increasing sample size while predictive performance remained comparable to the correctly specified Cox benchmark. In the Steno application, the Poisson Super Learner remained close to the best-performing individual learner or benchmark across prediction horizons, whereas the Discrete Poisson Super Learner performed less well. This illustrates the advantage of combining candidate learners when their relative performance varies across causes or prediction horizons. 

In the present implementation, the meta-learner coefficients do not depend on time. A time-varying meta learner could be obtained by allowing \(\alpha_{j,k}^s\) to depend on \(k\), at the cost of additional parameters and potentially greater instability in intervals with few events. Further extensions include adding more flexible learners to the framework, such as the Highly Adaptive Lasso or eXtreme Gradient Boosting.
\section*{Acknowledgments}
Approval for access to and use of personal data was obtained from the Capital Region of Denmark's Data Protection Office (approval no. P-2019-511). Approval for access to electronic patient records was obtained from the Danish Patient Safety Authority (3-3013-2959/1). According to Danish legislation, ethical committee approval and informed consent were not required for this register-based study.

During the preparation of this work, the authors used ChatGPT to correct typographical errors and assist with coding. The authors reviewed and curated the tool's output and take full responsibility for the content of the manuscript.

Gabriele Pittarello was funded by Novo Nordisk Foundation grant NNF23OC0084961.

\paragraph{Conflict of Interest}
None declared.

\section*{Data Availability Statement}
The data underlying this article cannot be shared publicly because doing so could compromise the privacy of the individuals who participated in the study. A synthetic version of the data can be generated using the \texttt{simulateStenoT1} function in the R package \texttt{poissonsuperlearner}.

\bibliographystyle{plainnat}
\bibliography{bibliography}

\section*{Supplementary Material}
Web Appendices, Web Tables, and Web Figures referenced in Sections 3, 6 and 7, and replication code covering the simulation study in Section 7, are available from the GitHub repository at \url{https://github.com/gpitt71/replication-material-psl} and will be made available on the journal website in case of acceptance. The simulation workflow was implemented in R \citep{Rreference} using the \texttt{targets} package \citep{targets}. The \texttt{poissonsuperlearner} R package, version \texttt{0.2.0}, used in the analyses and to generate synthetic data, is available from CRAN at \url{https://CRAN.R-project.org/package=poissonsuperlearner} \citep{poissonsuperlearner}.

\end{document}